# Design and Implementation of a Dual Uncrewed Surface Vessel Platform for Bathymetry Research under High-flow Conditions

**DINESH KUMAR[1], AMIN GHORBANPOUR[1], KIN YEN[1], AND IMAN SOLTANI[1](Member, IEEE)**

[1]Department of Mechanical and Aerospace Engineering, University of California - Davis, Davis, CA 95616 USA

Corresponding author: Iman Soltani (e-mail: isoltani@ucdavis.edu).

https://github.com/Soltanilara/Twin-USV

**ABSTRACT** Bathymetry, the study of underwater depth and topography, relies on detailed sonar mapping of the underwater structures. These measurements, critical for infrastructure health monitoring and hazard detection, often require prohibitively expensive sensory equipment and stable measurement conditions. The high financial risk associated with sensor damage or vessel loss creates a reluctance to deploy uncrewed surface vessels (USVs) for bathymetry. However, the alternative, crewed-boat bathymetry operations, are costly, pose significant hazards to personnel, and frequently fail to achieve the highly stable conditions necessary for bathymetry data collection, especially under challenging conditions such as high currents. Consequently, further research is essential to advance autonomous control, navigation, and data processing technologies, with a particular focus on bathymetry while ensuring safety under extreme conditions. There is a notable lack of accessible hardware platforms that allow for integrated research in both bathymetry-focused autonomous control and navigation, as well as data evaluation and processing. This paper addresses this gap by detailing the design and implementation of two complementary (dual) research USV systems tailored for uncrewed bathymetry research. This includes a low-cost USV for **N**avigation **A**nd **C**ontrol research (NAC-USV) and a second, high-end USV equipped with a high-resolution multi-beam sonar and the associated hardware for **B**athymetry data quality **E**valuation and **P**ost-processing research (BEP-USV). The NAC-USV facilitates the investigation of autonomous, fail-safe navigation and control technologies, emphasizing the stability requirements for high-quality bathymetry data collection while minimizing the risk to expensive equipment, allowing for seamless transfer of validated controls to the BEP-USV. The BEP-USV, which mirrors the NAC-USV hardware, is then used for additional control validation and in-depth exploration of bathymetry data evaluation and post-processing methodologies. We detail the design and implementation of both USV systems, open source hardware and software design, and the bill of material. Furthermore, we demonstrate the system's effectiveness in both research and bathymetric applications across a range of operational scenarios. All the information are available at: https://github.com/Soltanilara/Twin-USV/.

## I. INTRODUCTION

Bathymetry, the study of underwater depth and topography, is essential for numerous applications, including infrastructure health monitoring [1], preventive maintenance [2], navigation safety [3], underwater archaeology [4], and scouring evaluation [5] to name a few. Some applications related to underwater engineering [6], marine sciences [7], archaeology [8], and defense [9] require high-resolution bathymetric surveys.

A safety critical example application of bathymetry is related to the inspection of underwater infrastructure, particularly bridges [1], [10], [11]. This involves assessing the structural integrity and safety of submerged infrastructure such as bridge foundations, pipelines, and cables. Such inspections are vital for identifying potential issues that could lead to catastrophic failures. For underwater infrastructure as an example, regular bathymetry inspections seek to detect signs of damage or deterioration including scour (the erosion of sediment caused by flowing water around bridge foundations), as well as cracks, corrosion, and deformation. Scour, particularly, poses a substantial threat by potentially undermining structural stability and leading to disastrous bridge collapses if unaddressed (Fig. 1). A notable incident underscoring the importance of frequent bathymetry was the collapse of the Schoharie Creek Bridge in New York in 1987, which resulted in ten fatalities [12]. This collapse was triggered by severe scour due to snow melting and record rainfall that quickly eroded the sediment beneath a bridge pier.

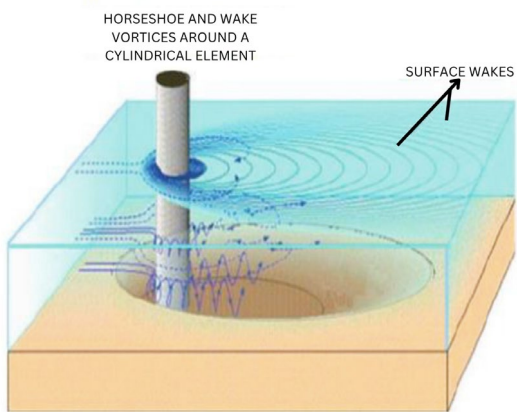

HORSESHOE AND WAKE VORTICES AROUND A CYLINDRICAL ELEMENT

SURFACE WAKES

**FIGURE 1.** Scour holes around a bridge pier [13].

High-precision surveys required in such safety-critical applications necessitate sophisticated sensory equipment, which can be prohibitively expensive. These systems include MultiBeam EchoSounders (MBES) that emit thousands of sound pulses per second across a wide area of the sea/riverbed and capture their echoes. They further include Differential GPS (DGPS) or Real-Time Kinematic (RTK) GPS as well as high precision IMU to allow for subsequent high-precision registration of measurements. The financial risk associated with potential damage or loss of the bathymetry equipment is significant. This often creates a tendency to adopt crewed operations, despite the associated hazards, discomfort, labor intensity, and high costs. As a result, the frequency and quality of data collection are ultimately compromised. Additionally, many locations with elevated scouring risk, such as river narrowings under bridges, feature persistent high flow conditions. These increase the risk to human operators and complicate the acquisition of reliable, motion-artifact-free data. Given these challenges, there is a critical need to advance research and development of robust, fail-safe bathymetry USV systems designed to collect high-quality data in diverse and demanding environments, including areas with high-flow conditions.

The first step in automated bathymetry research is developing hardware platforms that enable safe evaluation of new techniques. These platforms must support testing of bathymetry USV navigational and functional robustness, as well as high-quality data collection and processing under real-world conditions, without excessive concern about potential equipment risks. Such a research platform must be equipped with fail-safe mechanisms to prevent unpredictable or risky actions that could compromise hardware, yet still offer researchers the freedom to explore novel control, instrumentation, machine learning and design approaches. This balance is inherently challenging, as the flexibility required for experimental research often conflicts with the need for strict monitoring and constraints that enhance safety. Currently, there is a gap in the availability of such research platforms that can meet these conflicting requirements. This paper aims to bridge this gap by detailing the design and implementation of a dual-USV platform specifically tailored for bathymetry research.

We introduce implementation details of a dual-USV system, designed for autonomous bathymetry research. Our platform comprises an NAC-USV and a BEP-USV system, engineered to operate across a spectrum of safety and flexibility settings. The NAC-USV facilitates navigation and control development and testing, offering capabilities ranging from direct propeller control to high-level navigation including position and speed/heading control. The BEP-USV facilitates the implementation and further evaluation of methods tested on the NAC-USV, with additional focus on bathymetry-specific requirements under more extreme environmental conditions. The BEP-USV also serves as a platform for investigating techniques to evaluate and process bathymetry data collected under real-world conditions. By separating the NAC and BEP platforms while maintaining identical hardware specifications and implementation requirements, we mitigate concerns about safety, thereby removing barriers to flexible research.

This paper provides detailed descriptions of the hardware and software implementations for both USV systems, lists all components, and shares the CAD models and software publicly to encourage transparency and collaboration within the research community. The structure of this paper is organized as follows: Section III outlines the bathymetry mission requirements. Section IV explores the mechanical design of the dual-USV system, including methodologies for drag and thruster calculations. Details of the mechanical and electrical hardware implementations are covered in Sections V and VI, respectively. Section VII describes the control strategies employed in the dual-USV system. Finally, Sections VIII and IX present the experimental results and conclusions, respectively.

## II. PRIOR WORK

While no prior work has directly addressed the specific challenges of bathymetry under high-flow conditions, several studies from overlapping domains offer valuable insights. These related efforts can be broadly categorized into four key areas: (1) standardization of data collection and evaluation,

(2) USV control and navigation, (3) bathymetry data post-processing, and (4) USV hardware development. Collectively, these works contribute foundational knowledge that can be adapted to address the unique demands of high-flow bathymetric surveying.

**Bathymetric Data Quality and Survey Standardization:** Part of past research has attempted to provide an understanding of the effect of environmental disturbances on the quality of bathymetry data and to support standardization of data collection under diverse conditions. To be universally applicable, bathymetric measurements must adhere to established hydrographic standards. This includes ensuring that data collection and processing methodologies, as well as survey accuracy, are in alignment with international standards such as the IHO S-44 Special Order [14]. The quality of data is affected by several factors, including the motion of the system hosting the sonar [14]. Bathymetric surveys conducted in challenging environments, ranging from coastal regions to flood scenarios and areas with persistent high flow conditions, such as river narrowing under bridges, often experience significant vessel motions. In conditions with high winds and waves, due to the induced movements, small inaccuracies in sensor calibration can lead to large accumulated errors including inaccurate measurements of depth [15]. Due to USVs' smaller mass when compared to crewed vessels, these effects may be more pronounced, thus adding to the challenges of data collection. [14] highlights how the motion of USVs in different sea states can drastically impact data quality, showing a marked decline in data reliability at early sea state 3 (characterized by waves around 3 feet high and wind speed of 14 knots [16]). Ship-based surveys, with their larger mass and higher inertia, can typically operate in late sea states 3 (characterized by waves around 6 feet high, and wind speeds of 18 knots [16]) before data quality becomes compromised. While post-processing motion compensation techniques are adopted in advanced bathymetric systems, their effectiveness drops as vessel movements increase. These methods further face limitations caused by water clarity, surface currents, limited scan range and finally excessive cost [15].

**USV Control and Navigation:** Control and navigation play a crucial role in stable bathymetry data collection. Conventionally, for effective navigation and control of USVs, a mathematical model of the vehicle dynamics is needed. Fully capturing the dynamics of a USV is inherently complex. In addition to the standard six degrees of freedom, the system is subject to highly nonlinear hydrodynamic effects arising from continuous interaction with the water, further increasing its nonlinearity and overall complexity [17]. As such, more simple models such as those focusing on 2D plane motion involving 3 DoF (surge, sway, and yaw motions) has been used [18]. In addition to physics-based modeling, black-box modeling [19] uses data-driven techniques like machine learning to predict behavior without detailed physical models, relying heavily on data quality and quantity. Once a suitable system model has been identified, motion planning and motion con-trol is the logical next step. Motion planning is essential for accurate and efficient bathymetry surveys, enabling USVs to navigate complex aquatic environments while systematically covering the survey area. A well-designed motion planning system optimizes the coverage path, ensuring high-resolution depth data collection with minimal redundancy or gaps. Given the presence of dynamic obstacles, varying water depths, and environmental disturbances such as currents and winds, the motion planner must generate safe and collision-free paths while adhering to survey constraints.

Various native motion planning algorithms are available, each suited to different operational conditions. Geometric-based methods, such as the Line of Sight (LOS) and Integral Line of Sight (ILOS) algorithms [20], assume a direct path between the current position and the destination. While they are computationally efficient, they are also highly susceptible to environmental disturbances. Graph-based algorithms, such as A* and D*-Lite, provide optimal paths through known environments, whereas sampling-based methods, like Rapidly Exploring Random Trees (RRT) or RRT*, are used in complex and dynamic environments [21], [22]. Improvements in these algorithms include addressing issues with A* such as limited neighborhood search, unconstrained steering direction with an improved A* algorithm [23] or adapting the RRT algorithm for obstacle avoidance [24]. However, motion planning alone does not guarantee precise execution; this is where motion control comes into play. Motion control refers to the real-time execution of the planned trajectory using low-level control strategies. Adaptive controllers, such as the L1 Controller [25] fall into this category. There are also methods that integrate path planning and control such as Model Predictive Control (MPC) [26], which offer real-time adjustments to compensate for disturbances, making them well-suited for marine applications. The choice of motion planning and control strategy depends on mission objectives, environmental complexity, and computational constraints. More recently, advanced solutions, such as machine learning (ML)-based path planning and control, hold promise against environmental disturbances, as similar approaches have proven successful in UAV systems for gust rejection [27]–[29]. Some progress has already been made in this area [30], [31], but the developed methods are often tailored to specific marine conditions, limiting their generalization. Expanding these solutions to create adaptive, environment-agnostic control systems could greatly enhance the operational flexibility of USVs in bathymetry surveys. Furthermore, the high cost of MBES sensors (often in the hundreds of thousands of dollars) and USVs necessitates stringent safety measures and strong stability and performance guarantees. These precautions help prevent platform loss, sensor damage, and potential harm to other vessels and personnel. Although advances in collision avoidance and disturbance handling have been achieved in USVs [32]–[34] direct application to bathymetric USVs remains difficult due to the unique operational challenges and specific performance requirements of these vessels.

**Bathymetry Data Post-Processing:** Regardless of the ap-

plied navigation and control techniques, residual motion artifacts will inevitably persist in the data. Thus, post-processing of data is considered another crucial step in hydroacoustic measurements [16]. The complexity and nuances of post-processing schemes are heavily influenced by the challenges and limitations of USV control. Therefore, post-processing must be dynamically integrated with control strategies, calling for a concurrent investigation of both vessel control and data processing techniques, ensuring that advancements in one area can inform and enhance the other. However, many of the state-of-the-art data processing techniques have been adopted from other domains with no consideration of the unique characteristics of bathymetric data. As an example, due to nearly flat seabed topography, the occurrence of invalid loop closures is unavoidable in bathymetric SLAM [35]. Furthermore, continuous environmental disturbances make USVs particularly prone to exacerbated drifts in dead reckoning, resulting in degraded resolution of MBES data [36]. This is of additional concern when dealing with GPS signal loss, which is common below overpasses and near bridge columns. In bathymetry, small IMU drift can lead to large errors, rendering the data unusable. Although, utilizing higher-grade IMUs, such as those compliant with MIL-STD 810 or MIL-STD 461, can be helpful [37], they significantly increase the cost of the platform and still suffer from limitations. In recent years, the research emphasis on bathymetry-specific data processing has predominantly been on relatively simple steps such as improvements in data acquisition [38] and data cleaning methods [39]. Consequently, despite its critical importance, bathymetry data post-processing remains a relatively under-explored area [40]. Much like navigation and control, bathymetry post-processing stands to benefit from recent advancements in machine learning. Studies such as [36] explore deep-learning methods for loop closure in underwater bathymetry mapping, but these efforts remain limited and warrant further expansion. Given the direct effect of GNC (Guidance, Navigation, Control) performance on bathymetry data post-processing challenges and requirements, tackling both GNC and post-processing aspects simultaneously could accelerate progress in this field.

**Development of USV hardware:** Currently, there are limited resources available to researchers for developing USV platforms tailored to bathymetry and related GNC research. To the best of our knowledge, there is no open-source research platform that allows for simultaneous investigation of GNC and data processing with a focus on bathymetric applications. Various research papers document the development of USV platforms, covering control system design, mechanical structure, and GNC algorithms in a limited and isolated fashion [41]–[44]. Hence, there is a need for a comprehensive hardware and software setup that supports research in this domain. This work seeks to bridge this gap by introducing a fully open-source research platform.

## III. MISSION REQUIREMENT

The primary objective of the BEP-USV platform discussed herein is to conduct bathymetry surveys in diverse environments, including high-current conditions. This requirement is crucial, particularly in areas like those near bridges, where frequent bathymetric surveys are necessary due to strong currents and limited access, presenting significant operational challenges. To fulfill its mission objectives, the USV must meet several critical performance requirements. These include the capability to reach speeds approaching 7 knots, enabling operation under high flow conditions. The system must also have sufficient battery capacity to operate continuously for at least one hour, ensuring comprehensive data collection in a single deployment. In addition, given the required instrumentation including multi-beam sensor, onboard computational resources and battery support, the boat should also have a payload capacity of at least 50 lbs. Given the extremely high cost of the onboard equipment, the bathymetry survey must be executed safely, minimizing the risk of damage or loss. As such, it is crucial to conduct numerous tests during the research phase to ensure fail-safe performance of the USV under targeted extreme conditions. This necessity underscores the need for research flexibility as well as implementation robustness, two inherently competing attributes. For this purpose, we propose the adoption of a secondary USV system, NAC-USV, which features similar specifications as those of the BEP-USV but is significantly less expensive and relatively easy to set up. Its primary purpose is to evaluate the hardware and navigation/control software under various conditions before deployment on the BEP-USV which then allows for further research with more emphasis on the final task of bathymetry data collection and analysis. The BEP-USV and the NAC-USV setups are shown in Figs. 2 and 3.

In the following, we discuss our design approach and implementation details. The design discussions apply to both boats and are not specific to either unless explicitly stated otherwise. It's important to note that the outlined requirements can be competing; for instance, enhancing payload capacity adversely affects both speed and endurance, making the design an iterative process. In the following, we discuss the mechanical design process in detail.

## IV. MECHANICAL DESIGN

We employ an iterative design process, beginning with the selection of a hull best suited to meet the initial design specifications. This is followed by detailed drag calculations to assess hydrodynamic resistance at the desired speed for the selected hull, which informs our estimates of the required thrust. The thrust estimate then guides our thruster selection. If the design proves infeasible—for instance, if the thrust estimate is excessively high—we refine the process, starting again with hull selection and continuing until all criteria are met.

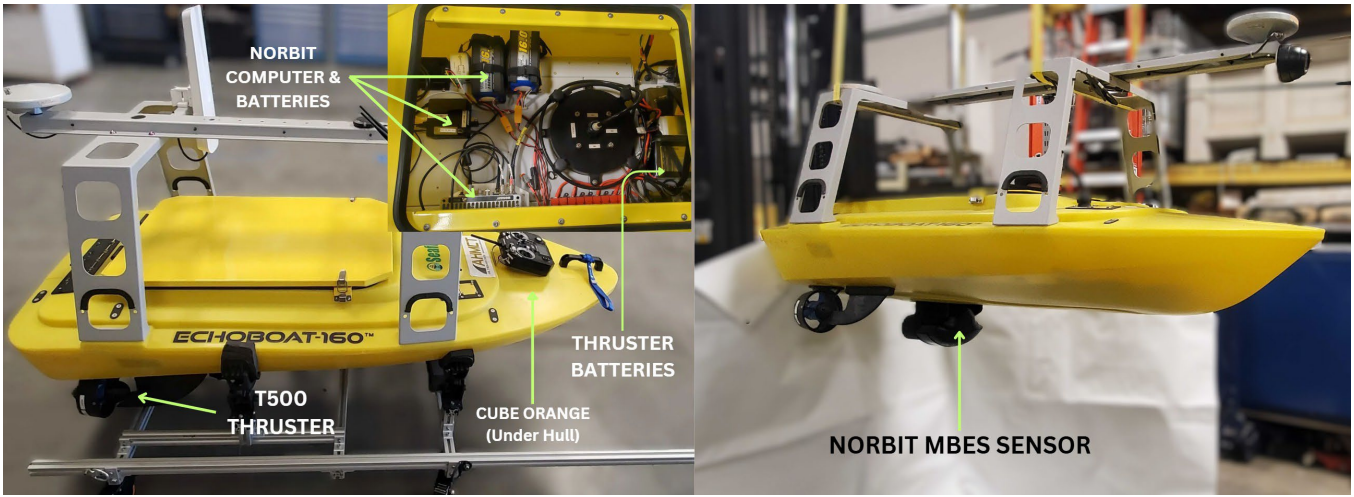

**FIGURE 2.** The BEP-USV.

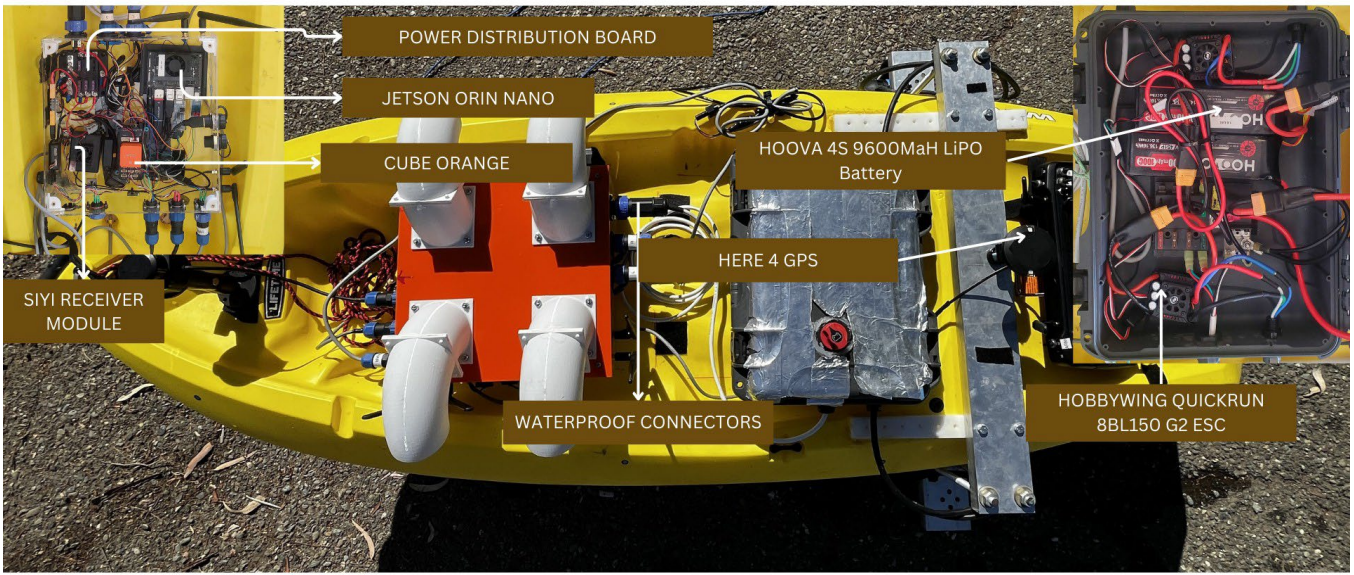

**FIGURE 3.** The NAC-USV.

### A. HULL SELECTION

The hull of the USV serves as its physical framework, providing structural integrity and buoyancy. It is typically made of durable, light-weight materials such as carbon fiber [45]. The hull profile, directly tied to USV hydrodynamics, plays a crucial role in the USV's stability, speed, maneuverability, and robustness to diverse environmental conditions [45]. Additionally, the hull is responsible to house and protect the internal components of the USV, such as the electronics.

There are two choices for the USV body design: catamarans and monohulls, as shown in Fig. 4. Unlike a monohull, a catamaran features a twin-hull design connected by a deck, which can potentially offer superior roll stability [45]. However, the deck space connecting the hulls is often used to house components like electronics, leading to challenges in evenly distributing the weight. This can hinder the full realization of the catamaran's inherent stability advantage. Monohulls, on the other hand, are easier to maneuver in tight spaces and often excel at upstream navigation thanks to their deep keel design [45]. Its superior maneuverability in tight spaces, for example near areas with vegetation, or near bridges where access is limited, is an important advantage.

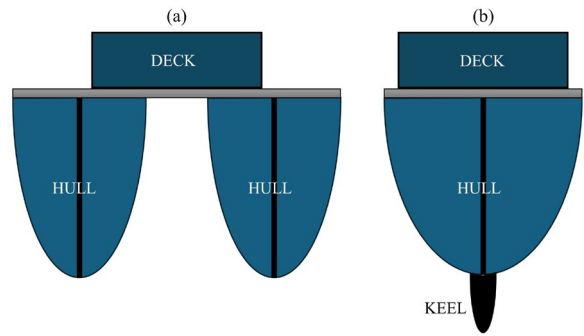

**FIGURE 4.** Types of USVs. (a) Catamaran, (b) Monohull

Additionally, the monohull's structure provides a simpler setup by allowing for easier integration and protection of all equipment inside the hull, reducing the need for extra wiring, and providing more flexibility in setting the center of gravity for stability purposes [45], [46]. They are also typically less expensive in terms of fabrication and maintenance compared to catamarans. As such, a monohull seems to be the better choice for bathymetry.

The hull shape also affects the stability, drag, and overall performance of a USV. Common hull shapes used for USVs include round-, rectangular-, V-shape-, and stepped-bottom designs [47] as shown in Fig. 5. Choosing the appropriate hull shape depends on the specific requirements of the USV, including its intended use, operational environment, and performance goals. Each shape offers distinct advantages and trade-offs in terms of stability, drag, manufacturability, and overall efficiency [48], [47]:

1) **Round:** Round hulls provide good buoyancy and stability and offer low drag at low speed and in calm waters. However, they may experience higher drag at high speeds and in rough waters, which can impact performance.

2) **Rectangular:** Rectangular hulls offer excellent stability and a large surface area for mounting equipment. However, they have higher drag compared to cylindrical and hydrodynamic shapes, making them less efficient in terms of speed.

3) **V-shape:** V-shaped hulls exhibit lower drag compared to rectangular and round hulls, and also provide good stability, especially at higher speeds. The main drawback is their complex manufacturing process, potentially increasing production costs.

4) **Stepped:** Stepped hulls feature one or more horizontal steps along the bottom of the hull, which reduce drag and enhance performance at higher speeds. These steps also improve the boat's wave-piercing abilities, making them suitable for rougher waters. The manufacturing process for stepped hulls is challenging and costly.

Among hull designs, V-shaped and stepped hulls are most commonly chosen for high-speed operations. The V-shaped hull is the most appropriate choice for our application as it offers an optimal balance of speed, stability, and performance [49].

### B. ESTIMATION OF TOTAL HULL RESISTANCE

To meet the maximum speed requirement, it is essential to accurately estimate the required thrust. However, this estimation must be preceded by determining the drag force associated with the selected hull, as it directly influences the thrust needed for the desired speed. In this section, we calculate the total hull resistance, $R_T$, which is a function of several parameters, including ship speed, hull geometry - such as draft - vertical distance between waterline to bottom of vessel's hull, beam - widest part of vessel, wetted surface area - surface area in contact with water, length, as well as water density and kinematic viscosity. The main components contributing to the total USV resistance are the fluid friction/viscous drag ($R_V$), wave drag ($R_W$) and air ($R_A$) resistance. The total resistance is written as [50], [51], $R_T = R_V + R_W + R_A$.

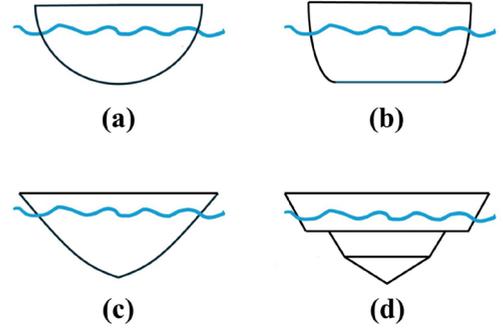

**FIGURE 5.** Common types of bottom hull design, a) round, b) rectangular, c) V-shape, d) stepped.

$R_A$ can be neglected ($R_A \approx 0$) since its contribution is only significant for large sizes, extremely high ground speeds, or strong wind conditions, which can largely be avoided in this application. Hence,

$$R_T = R_V + R_W \tag{1}$$

1) **Viscous Drag:** The viscous drag, $R_V$, increases with speed as shown in Fig. 6. It is caused by the friction and pressure forces on the hull due to the USV motion. Based on the assumption of a flat plate geometry for the hull, corrected for form factor as outlined in [51], the viscous drag can be approximated as:

$$R_V = \frac{1}{2}\rho V^2 C_F(1 + K)S_{wet} \tag{2}$$

where, $C_F$ is the friction coefficient (Eq. 3 [52]). The form factor $K$ (Eq. 4 [51]) accounts for the effect of hull geometry on viscous resistance. The wet surface area, $S_{wet}$, quantifies the submerged surface area (Eq. 5 [51]). We have:

$$C_F = \frac{0.075}{((\log_1 R_n) - 2)^2} \tag{3}$$

$$K \approx 19(\frac{\nabla}{L^2 \times D})^2 \tag{4}$$

$$S_{wet} = 2 \times (L \times B + B \times D + L \times D) \tag{5}$$

where $L$, $B$, $D$, $\rho$, $V$ denote the length, breadth, draft, fluid density and speed of the USV, respectively. $R_n = LV/\nu$ is the Reynolds number, and $\nu$ is the kinematic viscosity of fluid. $\nabla$ is the volume of fluid displaced by the boat.

2) **Wave Drag:** At higher velocities, wave drag may form a significant portion of the total drag as observed in Fig.

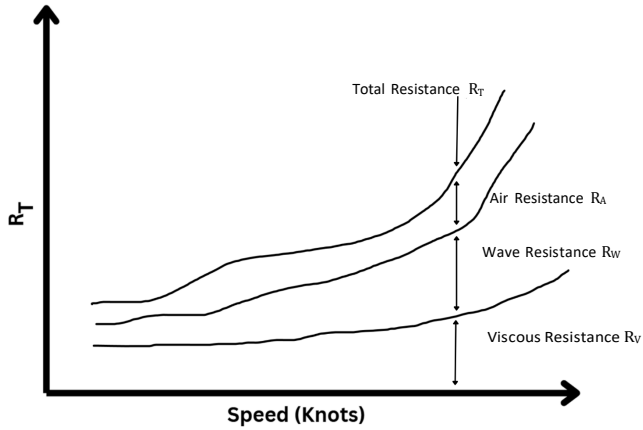

**FIGURE 6.** Viscous, wave, and air resistance are the key factors contributing to the total resistance encountered by a vessel moving through water [51].

6. Naval architects use the dimensionless Froude number, defined below, to predict the significance of wave drag [51]:

$$F_n = \frac{V}{\sqrt{gL}} \qquad (6)$$

where $g$ is the gravitational acceleration. It is shown that for small Froude numbers, $F_n < 0.3$, the USV's movement does not generate significant waves, and hence, the wave drag can be neglected [51]. For Froude numbers, $F_n > 0.5$ wave drag accounts for a significant part of the generated drag [51].

Estimating wave drag is challenging, as it requires complex experimental or numerical studies. [52] performed multiple resistance measurements on different vessels, and formulated a model for wave drag estimation:

$$R_W = \Delta \cdot c \cdot e^{m_1 F_n^{-0.9} + m_2 \cos(\lambda F_n^{-2})} \qquad (7)$$

where $\Delta$ is mass displacement($\nabla \times \rho_{water}$). In this work, the model parameters $c, m_1, m_2, \lambda$ are calculated per [52]:

$$c = 569 \cdot \frac{B}{L}^{2.984} \cdot C_M^{-0.7439} \cdot C_{WL}^{1.2655}$$

$$m_1 = -4.8507 \cdot \frac{B}{L} - 8.1768 C_p + 14.034 C_p^2 \\ -7.0682 C_p^3$$

$$m_2 = -0.4468 \cdot e^{-0.1 F_r^{-2}}$$

$$\lambda = 1.446 \cdot C_p - 0.03 \cdot \frac{L}{B}$$

where $C_p$, $C_M$, and $C_{WL}$ are dimensionless coefficients commonly used to characterize and compare marine vessels. For additional details, readers are encouraged to refer to [53]. These coefficients are calculated as follows:

$$C_P = \frac{\nabla}{A_M \cdot L} \qquad (8)$$

$$C_M = \frac{A_M}{B \cdot D} \qquad (9)$$

$$C_{WP} = \frac{A_{WP}}{L \cdot B} \qquad (10)$$

In Eqs. 8, and 9, $A_M$ denotes the submerged midsection area of the boat [53], as illustrated in Fig. 7. Similarly, in Eq. 10, $A_{WP}$ represents the surface area of the top-down view of hull at the waterline, as shown in Fig. 8.

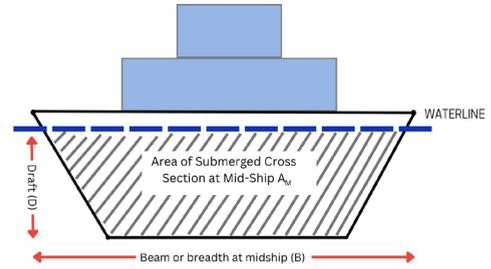

**FIGURE 7.** Area of submerged cross-section at midship ($A_M$) used to calculate the prismatic coefficient ($C_P$) [53].

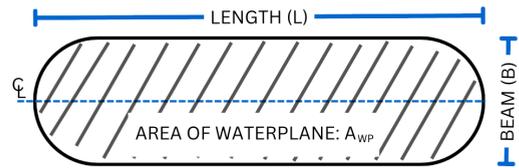

**FIGURE 8.** Area of Waterplane ($A_{WP}$) used to calculate the waterplane coefficient ($C_{WP}$) [53].

Given $R_V$ (Eq. 2), and $R_W$ (Eq. 7), the total resistance (Eq. 1) can be calculated.

### C. PROPULSION MECHANISM

Thrusters generate propulsion and steering, with propeller-based systems—such as open and jet propellers—being the most popular due to their versatility. For the bathymetry mission, open propellers are preferred over jet propellers due to their superior efficiency across a wide range of speeds and conditions [54]. They are also more cost-effective in terms of maintenance and occupy less space. Finally, unlike jet propulsion, open propellers can easily provide forward and reverse movements without the need for special mechanisms.

Depending on the desired steering mechanism, open propellers can be incorporated into a USV in various ways, including differential thrust, rudder-based systems (such as twin-screw setups), and vectored thrust. Differential thrust relies on thrust resultant and thrust imbalance across two or more propellers to respectively propel and maneuver the USV. This method eliminates the need for a rudder and allows for more effective control, especially for small to medium-sized vessels [55]. A twin-screw configuration includes one rudder behind each propeller. This setup is commonly used in larger vessels. In a vectored thrust system, the thrusters can rotate and change the direction of the resultant thrust force, enabling the USV to move forward, backward, and laterally, and also make sharp turns. This setup provides exceptional maneuverability, making it ideal for complex navigational tasks, although with additional complexity. Among these three steering methods, differential thrust presents a good trade-off between simplicity and performance and achieves acceptable maneuverability in most environmental conditions [56]. Hence, it is selected for our USV system.

### D. ESTIMATION OF THE REQUIRED THRUST

The first step in estimating the required active thrust, $T_a$, is to determine the total hull resistance, $R_T$, as was discussed in section IV-B. At a constant speed, the active thrust is equal to the total hull resistance, i.e. $T_a = R_T$. One also needs to account for moving thrust efficiency, $\eta_e < 1$, defined as the ratio between the active and the nominal propeller thrust $\eta_e = T_a/T_n$. Therefore, the nominal thrust of the propulsion system is higher:

$$T_n = \frac{T_a}{\eta_e}$$

$\eta_e$ is often provided in the data sheets of off-the-shelf propellers. $\eta_e$ is typically, $0.15 < \eta_e < 0.7$ [57].

Following the estimation of the required nominal thrust, $T_n$, including a safety margin, $k_s$, is advisable to account for the uncertainty of external factors, such as unexpected extra thrust requirements during aggressive maneuvering for collision avoidance or in adverse environmental conditions, e.g., when encountering stronger than expected currents or winds. The final estimate of the required thrust, $T_r = k_s T_n$, can inform the sizing and number of the propellers.

## V. MECHANICAL HARDWARE IMPLEMENTATION

Following the design process discussed earlier, in the following we go through each step and explain the implementation details for both BEP and NAC USVs.

### A. BEP-USV

We examined several commercial off-the-shelf (COTS) boats. Our evaluation criteria included maximum achievable top speed, maximum payload, and compatibility with commercially available propulsion systems. Additionally, we paid close attention to the equipment bay, giving priority to those specifically designed to reliably house critical components such as sensors, batteries, control systems, communication devices, and other necessary hardware for bathymetry USV's operation. Finally, to ensure the boat is easily portable for deployment, we limited the overall USV length to under 2 meters, making it compatible with a standard-bed pickup truck in the United States [58]. In case manual transportation is needed in any stage of the deployment, assuming the involvement of 2-3 operators, based on single person max load of 51lbs (23Kgs) [59], we attempt to keep the overall weight (Hull weight + payload) near or below 70 kg. Keeping the overall weight low also reduces the risk to other boats and individuals in case of collision.

The heaviest part of the payload in a bathymetry boat corresponds to the multibeam sensor and the batteries. Adding other necessary electronics and processing/communication hardware, we estimated an upper limit of 30 kg for our total payload. Taking into account the high flow rates typically observed in Sacramento and other California rivers—such as the American River at Chili Bar and North Fork Dam, where water velocities range from 3 to 5 knots [60], along with the standard bathymetry survey speeds of 2–5 knots [61]—a maximum operational speed of 7 knots (3.6 m/s) has been selected. This speed strikes an optimal balance between navigating strong currents and ensuring efficient data collection during surveys.

Following a thorough market search, we selected the Echoboat-160 from Seafloor Systems [62], as shown in Fig. 2. The Echoboat-160 has an empty weight of 50 kg and a maximum payload capacity of 27 kg. The boat comes with two thrusters and can achieve a top speed of 3 knots (for empty hull). The boat design allows replacing the thrusters to increase the maximum speed. As shown in Fig. 2 the USV features an enclosure for mounting additional equipment. Inside the equipment bay, there is a mounting plate designed for equipment installation, which sits on dampers to reduce vibration and enhance safety. The length ($L$), width ($B$) and height ($H$) of the boat hull are 1.7m, 0.8m, and 0.24m, respectively.

In the ship design literature, to accurately define the boat geometry, a multitude of parameters may be used to refer to distinct but oftentimes very similar quantities. For example, the length may be denoted as $L_{WL}$ (length at waterline) and $L_{BP}$ (length between perpendiculars). The width is often represented as $B$ (beam) and with considerations of the waterline as $B_{WL}$ (beam at waterline). Similarly, draft is denoted as $T$ (draft) and $T_{AVG}$ (average draft). For simplicity, we assume that each of these parameters can be represented with a single quantity and hence, use $L$, $B$, and $D$ to represent length, width and draft, as provided by the boat manufacturer. Given the dimensions and geometry of the Echoboat-160, and the maximum allowed payload of 27 kg, we adopt the quantities listed in table V-A for drag and thrust estimation.

For the desired speed, per Eq. 6 the Froude number is $F_n = 0.85$. Hence, both wave and viscous drag will contribute to the total drag. Following the procedure discussed in section IV-B, we calculate the required coefficients, which are provided in table 1. The drag estimation results are summarized in table 2 indicating a total drag of $R_T = 215.23 N$. Given the moving thruster efficiency of $\eta_e = 0.5$ (see section VI-4) we have $T_n = 430.46 N$, and assuming a safety factor of $k_s = 1.25$, the final required thrust, $T_r = 538.07$.

### B. NAC-USV

For the NAC-USV, our goal is to select a cost-effective, lightweight, and durable off-the-shelf small boat that is simi-lar to the BEP-USV in terms of the hull shape and size. After reviewing the options available on the market, we opted for a generic kayak, widely available through various vendors at a low cost. The selected kayak (Lifetime Wave) measures 1.82m (L), 0.61m (W) and 0.2m (H), weighs 8.2 kilograms, and has a weight capacity of 59 kilograms. The dimensions match the BEP-USV and allow for considerable flexibility for installing electronics and adjusting mass distribution to align the center of mass with that of the BEP-USV. For any initial tests in the research phase of the control and navigation, the kayak may only carry essential electronics weighing approximately 12 kilograms i.e. for a total weight of nearly 20 kg. In this phase maintaining a low overall weight is essential to minimize the risk associated with potential collisions and loss of control. Following successful preliminary evaluations, the weight of the boat can be incrementally increased up to 67.2 kilograms using dummy masses to closely approximate that of the BEP-USV and extend control evaluations.

| Parameter | Value |
|---|---|
| Max Velocity, $V$ | 7 knots = 3.6 m/s |
| USV Length, $L$ | 1.7 m |
| USV Width, $B$ | 0.8 m |
| Draft, $D$ | 0.25 m |
| Max Mass, $M$ | 77 kg |
| Midsection Area | 0.231 m² |
| Waterplane Area | 1.075 m² |
| Water Kinematic Viscosity, $v$ | $1.002 \times 10^{-6}$ m²/s |
| Water Density, $\rho$ | 1000 kg/m³ |
| Volume Displacement, $\nabla$ | 0.077 m³ |

**TABLE 1.** Coefficients and parameters for drag estimation for BEP-USV

| Parameter | Value |
|---|---|
| Wet Surface Area | 3.32 m² |
| Reynolds Number, Re | 5938123.75 |
| Midship Section Coefficient, $C_M$ | 0.52 |
| Prismatic Coefficient, $C_P$ | 0.17 |
| Form Factor, $K$ | 0.9 |
| Waterplane Area Coefficient, $C_{WP}$ | 0.7902 |
| Friction Coefficient, $C_F$ | 0.00329126 |

**TABLE 2.** Estimated drag values for BEP-USV

| Parameter | Value |
|---|---|
| Viscous Drag, $R_V$ | 129.05 N |
| Wave Drag, $R_W$ | 86.18 N |
| Total Drag, $R_T$ | 215.23 N |

In the following, we discuss the electronics, control, and computing hardware, which are identical for the BEP and NAC USVs. Additional implementation details can be found on the GitHub page associated with this manuscript [63].

## VI. ELECTRICAL HARDWARE IMPLEMENTATION

### 1) Power Source

There are several options available for the main power source of the USV, including Lithium Polymer (LiPo), Lithium-Ion, Nickel-Metal Hydride, and Lead-Acid batteries. LiPo was chosen as the power source due to its several advantages. It is lightweight, yet offers high capacity and energy density. The high discharge rate ensures quick power delivery to the thrusters [64]. Given the power requirements as well as the desired operation time (at least 1 hour), we selected the 6S, 22,000 mAh, 25C battery from Tattu and the - 4S, 9600mAh, 130C battery from Hoova, for the BEP and the NAC USV, respectively.

### 2) Instrumentation and Control

At the heart of our dual-USV platform, we adopt Cube Orange, which is originally developed for the UAV industry. This embedded system readily includes some of the required USV sensors, namely the Inertial Measurement Unit (IMU), barometers, and compasses, as well as access ports for CAN bus, serial ports, and I2C. The Cube Orange is preferred over its alternatives, such as Pixhawk 2.4.8, due to its redundant IMUs, and more number of serial, and CAN bus ports. It also allows for the addition of other sensors such as range finders, and companion computers for more complex processing tasks. Cube-Orange supports multiple GPS modules - serial as well as CAN-GPS modules such as Here 4 Multiband RTK, which is the preferred GPS for this project owing to its accuracy and multiband GNSS - enhancing much needed reliability near buildings and trees. The GPS also uses the CAN bus port, hence leaving the existing serial ports available for other sensors. The Cube Orange can operate firmware such as Ardupilot [65] or PX4 [66], both of which support actuator control for vehicle navigation. For our dual-USV platforms, we opted for the Ardupilot firmware due to our prior experience with it and its in-built obstacle avoidance capabilities.

### 3) Ground Station and Telemetry

To steer the USV and monitor its telemetry and health, a ground station is typically used. This comprises a software component running on a computer or tablet connected to the USV via radio at commonly available frequencies of 930MHz, 2.4GHz, or 5GHz. The latter frequencies are often used when streaming live camera feeds from the USV. The hardware further includes a remote controller. The software interface receives and displays telemetry information and camera feeds. Radio communications are generally limited to Line of Sight (LOS), although advanced systems may include signal repeaters placed in the field to ensure reliable communication with the USV. Repeaters are particularly beneficial in surveying operations around bridge columns, where direct LOS may be obstructed, enhancing the mission's safety. We chose the SIYI MK32-HM30 Combo, a handheld ground station that integrates control and telemetry into a single Android tablet, has a long transmission range of 15 kilometers, dual camera feed capability, and is compatible with signal repeaters. On the software side of the ground station, two popular options are Mission Planner [67] and QGroundControl [68]. QGroundControl is a better choice for research due to its cross-platform support, built-in video streaming support and user-friendly interface.

### 4) Speed Controllers and Number of Propellers

The speed controller, also known as Electronic Speed Controller (ESC) receives PWM signals from the Cube-Orange and regulates the voltage input to the propeller. Based on our thrust estimation (section V-A) we chose the Blue Robotics T500 propeller, along with the Hobbywing Quicrun WP 8BL150 G2 ESC. In general ESCs are chosen based on voltage, and current requirements of the thruster. It is important to ensure the selected ESC is capable of bidirectional motor control. As per T500 datasheet [69] the thruster has a max static (nominal) thrust of 16.44Kgf (161N) at 24V. When in motion, a $\eta_c = 0.5$ reduction is expected [70]. As noted before, we include a safety factor $k_s = 1.25$, leading to a final thrust of $T_f = 538.07N$ (see section V-A). To achieve this, four T500 thrusters are needed.

### 5) Companion computer

The Cube Orange is only compatible with simple sensory modalities such as range finders and support simple controllers such as PID. As a research platform, the dual-USV system should be able to accommodate more advanced sensors with higher data throughput, such as cameras and 3D lidars. It should also allow for the implementation of sophisticated control algorithms, image processing, machine learning, and other advanced computational tasks. For this purpose, the system is augmented with a companion computer. The companion computer can simultaneously access the Cube Orange to utilize its sensors (both embedded and external) while also interfacing with additional advanced sensors like cameras and lidars. Depending on the research requirements, the companion computer can either directly consume its collected sensory data e.g. through machine learning, or after additional processing and sensor fusion to extract ego-motion or obstacle position/speed information, can relay the results to the Cube Orange for processing by its embedded controller. Furthermore, information from either Cube Orange or the companion computer may be relayed to the user for research and monitoring purposes. This integration allows us to leverage the companion computer's flexibility and enhanced functionality on the USV while also benefiting from the fail-safe and already established, community-tested capabilities of the Cube Orange.

Given the potentials of machine learning in the control of autonomous mobile systems, we selected the Nvidia Jetson series for the companion computer due to its reasonably powerful GPU, suitable for real-time computer vision and other machine learning algorithms, and its energy-efficient design for embedded applications.

### 6) Sensors: Obstacle Detection and Range Estimation

For obstacle avoidance, sonar, and single-beam lidar can be natively connected to the Cube-Orange. Ardupilot supports a range of such sensors, and includes a few community-evaluated path-planning and obstacle-avoidance algorithms. As noted earlier, in addition to sensors that are natively supported on the Cube Orange, through the use of a companion computer more complex sensors such as stereo cameras or multibeam lidars can be used to provide 3D point cloud data, and detect and classify obstacles, etc. The companion computer can read/process more complex sensory data, and encapsulate processed results into Micro Air Vehicle Link (MAVLink) messages [71], which can then be received and processed by ArduPilot on the Cube Orange. Our proposed design uses the following sensors:

**1. Blue Robotics Ping 2 Sonar sensor**: This is an underwater sonar sensor, developed by Blue Robotics. Before integrating the sonar sensor into the USV, we conducted preliminary tests in a UC Davis pool. By collecting sensory data at different sensor angles, we evaluated its reliability and investigated how the mounting angle affects ground plane detection versus the identification of forward obstacles. A 15-degree mounting angle (with respect to the water surface) was found to provide a good balance, effectively detecting forward obstacles while also identifying shallow conditions that could cause the boat to become grounded in mud or collide with the waterbed which can damage the MBES. An angle larger than 50° was found to be more suitable for ground-plane detection only. This sensor could be directly integrated with the cube orange. However, for sensor fusion when used in conjunction with other sensors, it is connected to the companion computer.

We encountered several challenges in integrating the 1-D sonar sensor, all related to reliable detection of underwater obstacles, such as riverbanks, slopes, and piers. Multiple field tests were conducted to evaluate and characterize the performance of the sonar. These tests were crucial in determining baseline confidence levels, adjusting sampling rates and allowable detection range, and finally developing digital filters for smoothing of the measurements. The sensor was found to be highly sensitive to water turbulence and hull-induced noise, both of which significantly impacted its ability to detect obstacles consistently. To address these issues, the sonar was mounted at various locations on the boat, with different levels of separation from the hull, to identify the best configuration that minimized interference. Extensive data, including sonar readings, confidence levels, and GPS coordinates, was collected and analyzed to better understand the sensor's behavior. Furthermore, moving average filters were found to be effective in reducing noise and stabilizing readings.

**2. Slamtec S3 Lidar**: This sensor is IP-rated and features an ambient light anti-interference capability of up to 80,000

lux [72], ensuring reliable performance even in high-light environments, making it a suitable candidate for USV applications. This lidar is not natively supported by Ardupilot, and hence requires additional processing on the companion computer before encapsulating into MAVLink messages to the Cube Orange. The lidar sensor proved to be significantly easier to characterize, as it consistently produced reliable data. Operating at a sampling rate of 32kHz, the lidar occasionally generated unreliable measurements; however, the average distance to obstacle over several consecutive measurements within each FoV(Field Of View) sector remained consistent, resulting in a stable and robust sensing. One significant challenge with this sensor was related to the required downsampling. The Slamtec S3 lidar produces up to 32,000 samples per second over a full 360-degree sweep. However, ArduPilot accepts only 72 discrete distance values, each representing a 5-degree FoV sector around the USV. After testing multiple downsampling strategies, the selected approach involved dividing the 360-degree scan into 72 bins, each spanning 5 degrees. Within each bin, a weighted average was computed, where the weights favored values closer to the minimum distance (with the highest confidence) in that sector. This method preserved the closest obstacle information while smoothing out noise, enabling robust integration with ArduPilot's obstacle avoidance framework.

The selected lidar and sonar sensors are connected to the companion computer, where a script fuses their corresponding measurements into a 72 element array capturing 360° around the USV. This array is then packaged into Obstacle Distance message type of MAVLink, which is published to ArduPilot via the serial port providing a unified input that simultaneously covers obstacles above and below the water and can be used by either the obstacle avoidance algorithms available on ArduPilot or those custom implemented on the companion computer. In our implementation, when using ArduPilot's native obstacle avoidance methods, the USV either slows down or comes to a complete stop depending on the obstacle's proximity to the USV.

### 7) MultiBeam Echo Sounder

The BEP-USV is instrumented with a Norbit iWBMSe Multibeam Echo Sounder (MBES). This is a compact, high-resolution, broadband multibeam sonar system designed for bathymetric surveying. Aside from the size requirements, the MBES selection was driven by our budget limitations. Priced > $120, 000, this MBES system is on the lower end of the typical $75k to $500k range for commercially available systems. It is equipped with an advanced GNSS-aided inertial navigation system, and hence, is capable of position and roll stabilization, which enhances accuracy. This sonar operates at a center frequency of 400 kHz, with a frequency range of 200 to 700 kHz, and a ping rate of up to 60 Hz. Additional details can be found in the sensor datasheet [73].

An overview of the USV system architecture, along with the major components, and communication protocols is demonstrated in Fig. 9.

## VII. FLEXIBLE CONTROL ARCHITECTURE FOR RESEARCH

In the rapidly evolving field of autonomous surface vessels, the ability to develop, implement, and test different control systems is crucial for driving innovation and advancing research. Research in this domain often demand varying levels of interaction with vehicle dynamics, from low-level thrust control to high-level strategic navigation. Hence, our design prioritizes flexibility to support a wide range of research objectives, enabling users to focus on specific aspects of marine autonomy without requiring deep expertise in every subsystem. This adaptability is particularly valuable in bathymetry, where requirements span a diverse range, from precise thrust control for obstacle avoidance and disturbance rejection to path planning for efficient data collection and environmental interaction. To address these needs, our USV platform supports three distinct control strategies. Each strategy progressively abstracts away the underlying dynamics, providing varying levels of control flexibility as detailed below.

### A. AVAILABLE CONTROL STRATEGIES

**1) Direct Propeller Control:** At the lowest level, our USV platform allows direct control of individual propeller speeds. This mode is particularly useful for studying the vessel's dynamics in detail, offering the fine-grained control needed for experiments focused on propulsion, hydrodynamics, and motion behaviors related to robustness, stability, maneuverability, and efficiency. This option bypasses the Ardupilot's all internal control loops where each thruster is linked to a separate radio channel, managed by the companion computer. It also permits remote control takeover from the ground station when necessary. This configuration lacks built-in safety measures and, therefore, must be used with caution. Further details on these safety considerations are elaborated in our GitHub repository [63].

**2) Velocity and Heading Control:** Raising the level of abstraction, this strategy enables researchers to control the USV's velocity and heading without delving into the complexities of hydrodynamics or the higher-order, or nonlinear dynamics of the vessel and its propellers. This intermediate level of control is useful for the development of system agnostic, and generalizable approaches to navigation. It suits applications such as machine learning, where algorithms like reinforcement learning dictate the direction and speed of the boat, focusing on strategic decision-making for example in the presence of dynamic and static obstacles and high-flow disturbances [74]. It is particularly useful in fields such as computer science, where it's assumed that low-level control specifics can be adapted and optimized for each platform by dedicated specialists. In our implementation, this control mode involves simple native PID loops of the Ardupilot firmware. In conjunction with L1 control [75], this mode can be used for constant trajectory

**3) Position Control:** The highest level of control abstracts all dynamics considerations, allowing the researcher

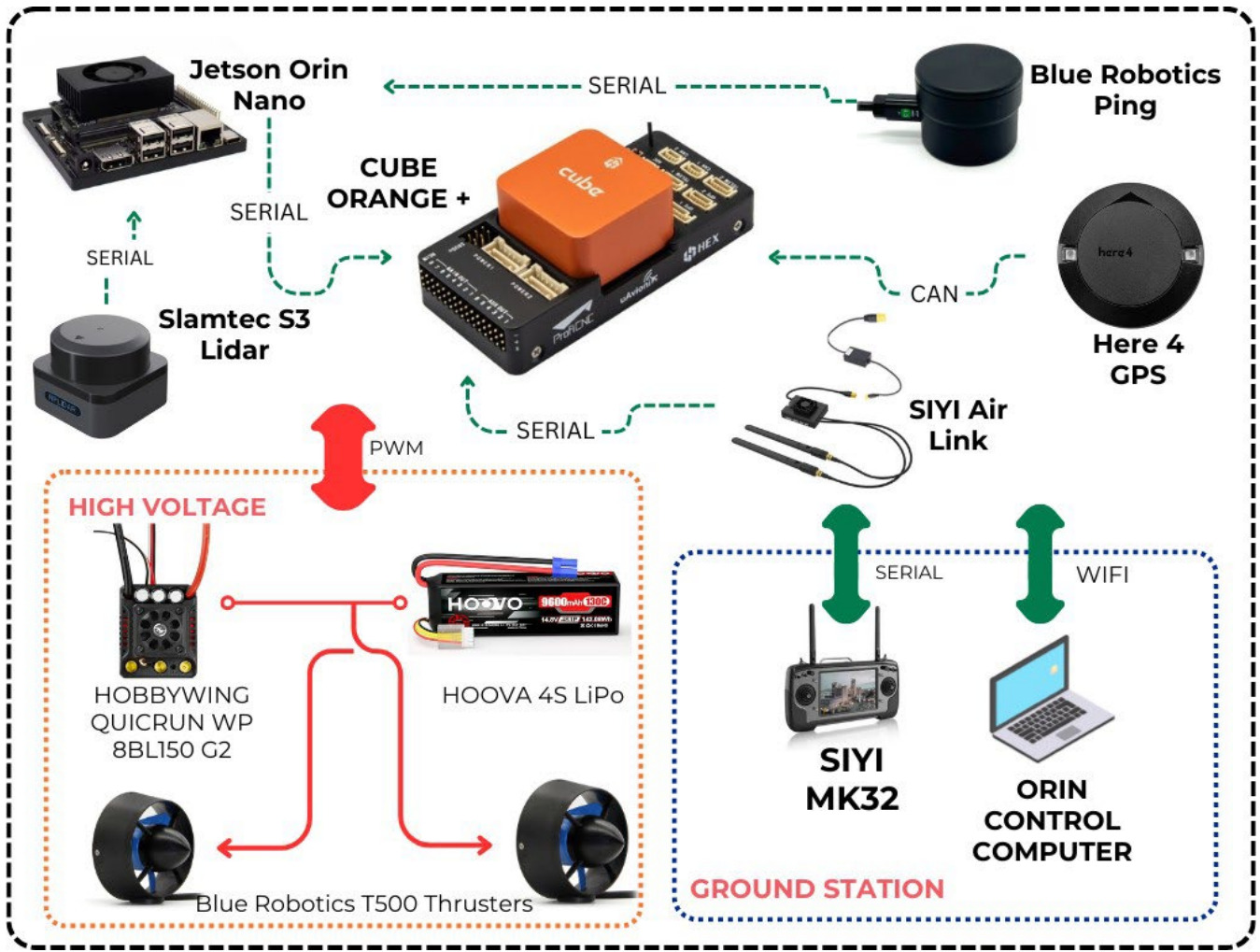

**FIGURE 9.** Electrical architecture of the USV and the associated components.

to focus solely on the USV's position. This strategy, which relies on the GPS measurements, is essential for studies that prioritize coarse path planning such as following zig-zag patterns across rivers for bathymetry scans, or static obstacle avoidance, where the primary concern is the geographic path of the vessel rather than the specifics of how it is achieved. By managing all underlying dynamics internally, this mode enables researchers to concentrate on tasks such as optimized data collection and processing.

### B. LIMITATIONS OF THE USV KINEMATICS, SENSING AND LOW-LEVEL CONTROL FRAMEWORKS UNDER STRATEGIES 2 AND 3

Even under a simplified 2D motion model with three degrees of freedom (surge, sway, and yaw), a USV equipped with only two thrusters is considered underactuated. For the low-level control of strategies 2 and 3, we employ ArduPilot's differential thrust steering mode, which directly controls surge and yaw, while sway remains unactuated, resulting in a non-holonomic system. The mathematical modeling of such systems is well established in the literature and omitted here for brevity; interested readers are referred to works cited in Section VII-C, such as [76] and [77]. In addition to the inherent kinematic limitations, both the BEP-USV and NAC-USV impose physical constraints, including limits on turning rate, speed, and thrust

ArduPilot relies extensively on onboard sensors such as IMUs, compasses, and GPS for guidance and navigation. However, these sensors are often subject to significant noise, drift, and low update rates, which directly impact USV control performance, ultimately affecting the quality of data collection. A common solution is to adopt higher-end sensor suites that offer improved accuracy and reliability. For example, this project utilizes the Here4 GPS, which provides centimeter-level accuracy [78], helping mitigate many of the issues associated with standard navigation hardware.

The default ArduPilot velocity and heading controller used in Strategy 2, as well as the trajectory position controller employed in Strategy 3, rely on closed-loop control methods that are agnostic to the underlying USV dynamics. While this model-free design improves generalizability and ease of use, it also introduces limitations that researchers should be aware of, particularly when precise control or dynamic responsiveness is required. In many practical scenarios, however, these default low-level controllers are sufficient, especially for operations conducted at relatively low speeds where the USV's dynamic constraints are not significant.

Heading/speed control (strategy 2) is handled using a combination of a *nested PID loop* for heading regulation and a separate *PID controller* for speed control. The heading control loop consists of two stages: an outer PID controller computes the desired yaw rate based on the heading error (i.e., the difference between the target and current heading), and an inner PID controller adjusts the differential thrust to achieve the commanded yaw rate. Independently, a speed PID controller regulates forward velocity by adjusting the total thrust applied to the vehicle. The outputs of these two control paths, *total thrust* from the speed controller and *differential thrust* from the yaw controller, are then combined to compute individual thrust commands for the left and right propellers. Specifically, the left thruster is commanded with $T_L = T_{\text{forward}} - T_{\text{yaw}}$, and the right thruster with $T_R = T_{\text{forward}} + T_{\text{yaw}}$. This allows the USV to simultaneously track a desired speed and heading. The position tracking under Strategy 3, uses an L1 guidance controller which adjusts the USV heading towards the waypoint while another PID controller regulates speed. [75]. Originally developed for fixed-wing aircraft, the L1 controller has been adapted for marine and ground vehicles due to its robustness and smooth path-following capabilities. The L1 controller computes a desired heading by calculating the turn rate required to converge to the given waypoint. The desired heading generated by the L1 controller is then passed to the nested heading control loop described earlier, which handles low-level actuation through differential thrust.

Researchers who require more advanced low-level control, such as machine learning-based, model-based nonlinear, or optimal control techniques, can readily replace our default controllers with custom implementations. More information about this can be found in the project GitHub page [63]. This modularity enables users to focus on their specific research goals, engaging with the vessel dynamics and low-level control only when necessary.

## C. IMPLEMENTATION OF THE CONTROL STRATEGIES ON ARDUPILOT

To implement the three control strategies discussed above, we use Ardupilot's available operation modes and further leverage two critical tools, the MAVLink messaging protocol and the ROS2. These tools are discussed in the following.

The ArduPilot's Manual mode to implement Strategy 1. In this mode, direct steering commands can be sent to the USV via the remote control (RC), providing full manual control to the operator. Leveraging this feature, each thruster was mapped to a dedicated RC channel, which was subsequently overridden using MAVLink messages from the companion computer to enable direct thruster control. Guided mode, used for Strategies 2 and 3, enables real-time navigation of the USV using commands from either the ground control station or a companion computer. Guided mode accepts various inputs such as position (latitude, longitude), speed, and heading, allowing the USV to move to specific locations, adjust its heading, or maintain a desired speed. Under Guided mode, if Obstacle Avoidance is enabled and properly configured, the USV will respond to detected obstacles accordingly. This allows researchers to implement Control Strategies 2 and 3 in conjunction with ArduPilot's native obstacle avoidance feature. ArduPilot also supports several other modes—including Loiter, Auto, Return to Launch (RTL), and Acro—which, although not used in our control strategies, provide additional flexibility and important fail-safe capabilities. Loiter mode holds the USV at its current position and heading, actively correcting for external disturbances such as wind or current. Auto mode allows the execution of a pre-defined route uploaded from the ground station, making it particularly suitable for conventional surveying missions and data collection. RTL mode autonomously returns the USV to its launch position, ensuring recovery in the event of mission termination or failure. Acro mode offers control over speed and turn rate, allowing the USV to maintain a desired heading while compensating for environmental factors, effectively balancing manual and automated control. In addition to these modes, ArduPilot incorporates critical fail-safe mechanisms, such as automatic RTL activation in the event of a low battery, and arming checks to ensure valid GPS and compass readings. These features are essential to achieving safe, reliable, and autonomous mission execution.

MAVLink [71], a lightweight and standardized communication protocol used in the ArduPilot firmware, facilitates the exchange of information between the USV, the ground station, and the companion computers. This protocol is essential for transmitting navigation control and supports a range of commands including velocity and location directives. The MAVLink protocol's capabilities are documented comprehensively in the official MAVLink repository [71].

In conjunction with MAVLink, we employ the Robotic Operating System 2 (ROS2-Humble) [79], which uses a publisher/subscriber model to facilitate component communication within robotic systems. In our setup, ROS2 is configured to publish MAVLink commands tailored to each of the three control strategies—direct thruster speed, velocity/heading, and trajectory control. Fig 10 outlines the ROS node architecture. It is noted that only one control strategy can be active at any given time. ROS2 publisher nodes awaits user value(velocity, position), while the corresponding subscriber node connects and packages these values as MAVLink messages before sending them to the ArduPilot over the serial bus. Researchers can modify the publisher nodes and integrate their code to tailor the system's functionality to their specific needs. Detailed instructions for connection to ArduPilot, code modifications, and setup are available in the GitHub Repo [63].

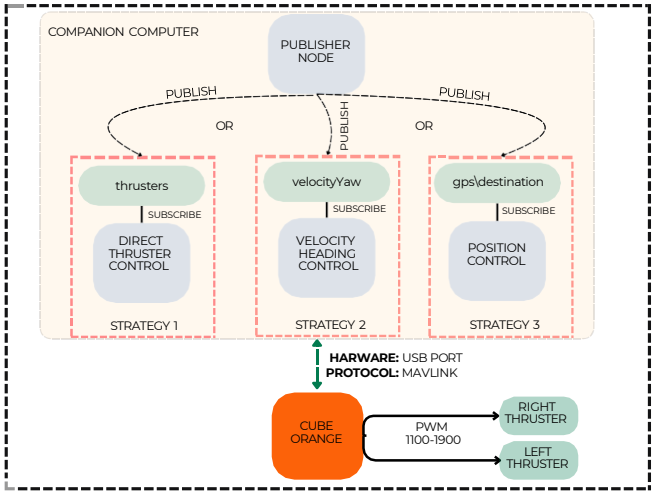

**FIGURE 10.** Use of ROS2 nodes and MAVLink with Ardupilot for control strategies 1-3

**TABLE 3.** Battery Specifications and Runtimes for NAC-USV and BEP-USV.

| Boat | Total Battery | Run Time | Top Speed |
|------|---------------|----------|-----------|
| BEP-USV | 22000 mAh − $6S \times 3$ | 3 Hours | 2.2 m/s |
| NAC-USV | 9600 mAh − $4S \times 1$ | 40 mins | 1.8 m/s |

## VIII. EXPERIMENTAL EVALUATION

The control strategies discussed in section VII were tested on field Fig. 11, speed, and heading were recorded onboard using ArduPilot's logging system. Table 3 lists the achieved top ground speeds for both NAC and BEP USVs along with their battery runtimes. At the time of testing, the river was relatively calm, so the ground speed and velocity relative to the water were expected to be similar. In our initial functionality tests, we used smaller, lower-voltage batteries on the NAC-USV than those installed on the BEP-USV VI-1 to simplify the NAC-USV setup. These batteries can be easily replaced as needed. Additionally, although our thrust estimation indicated a requirement for four T500 thrusters, we initially deployed only two propellers to validate system functionality and reliability before incorporating additional thrusters. Consequently, the maximum speeds recorded during these tests were lower than our target speed, particularly for the NAC-USV due to its smaller batteries. The BEP-USV reached a top speed of 2.2 m/s with just two propellers, demonstrating that as anticipated through thrust estimate, the desired speed (3.6 m/s) is attainable with the addition of two more propellers. The values reported for BEP-USV are taken during a bathymetry survey under non-flood conditions. Fig. 12 presents an example bathymetry scan collected on the BEP-USV from a bridge in California, illustrating a scour hole characterized by a sudden drop in underwater elevation.

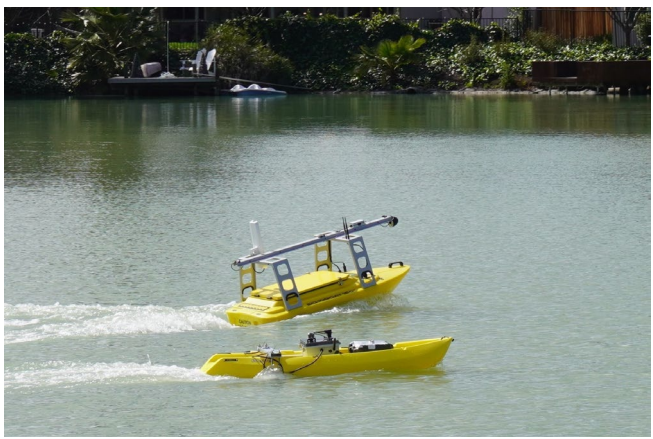

**FIGURE 11.** Field testing of NAC and BEP at Stonegate Village lake, Davis

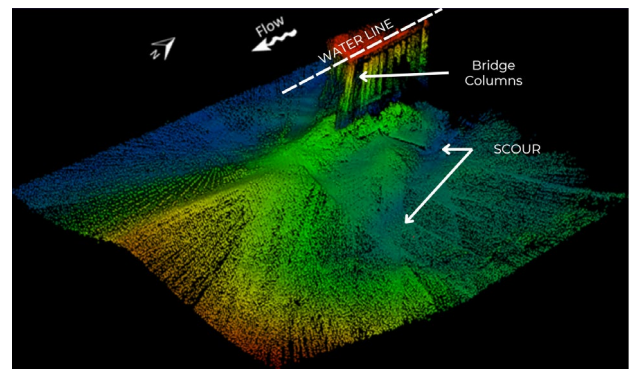

**FIGURE 12.** Bathymetry Data collected by the BEP-USV with scour hole detected. Survey conducted on the Sacramento river, Northern California

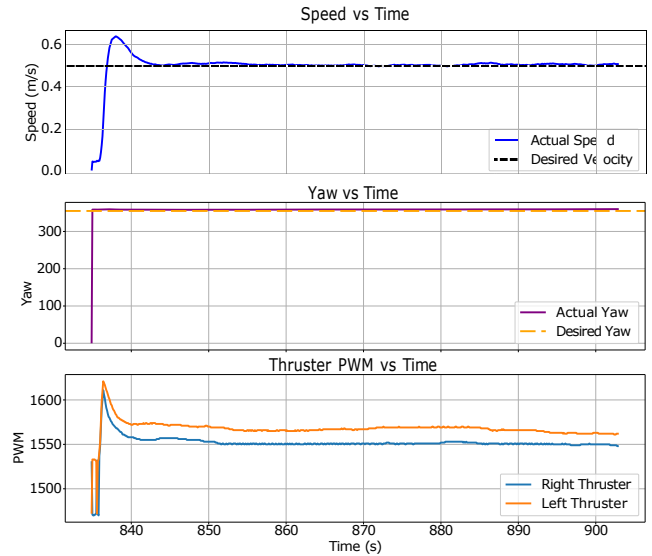

**FIGURE 13.** A speed and heading control test utilizing strategy 2 was conducted at Stonegate Village Lake Davis, California, during which the NAC-USV maintained a constant speed of 0.5m/s and a commanded heading of 355°

**Direct Propeller Control**: Tests on the NAC-USV demonstrated its ability to perform direct thruster control. While no quantifiable metrics are available for this control strategy, a video on the paper's GitHub page [63] qualitatively showcases its performance using a Logitech G920 remote controller [80]. Since the control hardware on the NAC and BEP USVs are virtual replicas, the experiment was conducted exclusively on the NAC-USV system.

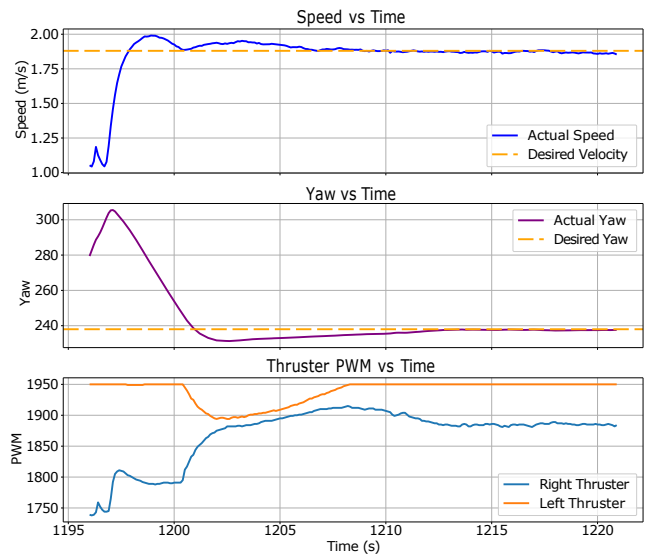

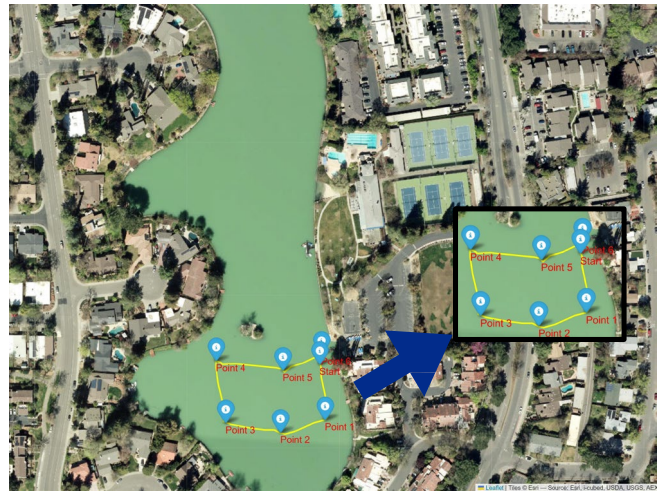

**FIGURE 15.** NAC-USV position control field test using strategy 3 at Stonegate Village Lake, Davis, California

**FIGURE 14.** A speed and heading control test utilizing strategy 2 was conducted at Stonegate Village Lake in Davis, California during which the NAC-USV maintained a constant speed of 1.8m/s and a commanded heading of 240°

**Velocity and Heading Control**: Fig. 13 and 14 show the NAC-USV's speed and heading response over time upon commanding the desired values. The USV achieves the commanded set points, thus validating that the system is capable of receiving and executing velocity and heading from the companion computer.This figure also includes the associated thrusters effort represented by the corresponding input PWM. As mentioned in section VII-C, this strategy utilizes cascaded PID loops within ArduPilot. Proper tuning of these controllers enhances both velocity regulation and trajectory tracking performance. There are two primary controllers that require tuning: 1) Yaw control, and 2) Speed Controller. [81] and [82] provide detailed instructions to help tune the USV. It is also possible to perform these tuning operations using Lua scripts as outlined in [83].

**Position Control**: Fig. 15 shows a sample executed multi-waypoint trajectory. The blue markers indicate the waypoints commanded by the user to the NAC-USV (in clockwise order starting from the bottom right), and the yellow line indicates the path followed by the NAC-USV. The trajectory is plotted over satellite imagery provided by Esri, Maxar, Earthstar Geographics, and the GIS User Community [84]. The node is set up such that upon reaching the target, the boat will enter into the loiter mode and attempt to stay within a 2-meter radius of the waypoint. This radius is a configurable parameter in Ardupilot. Waypoints can be overridden with a new waypoint at anytime to follow a coarse trajectory in real-time. Similar to the Velocity-Heading controller, proper tuning of the Position Controller in ArduPilot is essential for accurate navigation. Detailed instructions on the tuning process can be found in [85].

Multiple tests of the developed USV system under various environmental conditions revealed limitations in both the L1 Navigation Controller and the Cascaded PID Controller. In scenarios where trajectory inputs change rapidly, the control system tends to overshoot heading corrections or abruptly stop. In the presence of strong winds, the USV often halts to make dramatic heading adjustments before proceeding, leading to significant data quality degradation. While the cascaded PID Controller is simple and widely used, it is highly sensitive to environmental variations and requires frequent retuning for different conditions. Additionally, it is susceptible to system-level inconsistencies such as propeller imbalance, uneven weight distribution, and payload variations, all of which impact its performance. The smaller size of USVs, coupled with modeling inaccuracies, hardware degradation (e.g., worn-out motors, damaged propellers), and changing environmental conditions, presents inherent challenges in maintaining data quality as per IHO standards. Addressing these challenges requires the development and implementation of more advanced motion planning and motion control algorithms. For example, while the L1 controller can partially adapt to external disturbances better than traditional PD controllers [25], field tests showed that maintaining heading in high flow conditions remains problematic. Furthermore, these controllers lack adaptability in failure scenarios such as the loss of a propulsion motor, reducing their robustness in real-world operations.

## IX. CONCLUSION AND FUTURE WORK
This paper details the design and implementation of a dual USV system, namely the NAC-USV and BEP-USV, specifically tailored for bathymetry research under high-flow conditions. Both USVs were developed to meet the research community's needs, focusing on safe and adaptable approaches to control, navigation, data collection, and processing in uncrewed bathymetric operations.

The NAC-USV serves as a low-cost platform for developing and testing navigation and control technologies while minimizing the risk to expensive bathymetry equipment. The BEP-USV, a close replica of the NAC-USV with the addition of a high-resolution MBES, extends these tested capabilities to bathymetric surveying. As it mirrors the control system and hardware configuration of the NAC-USV, the BEP-USV can seamlessly integrate succesfully tested control and navigation frameworks directly from its NAC counterpart, enabling further research under safer and more predictable conditions.

As part of our system software architecture, we have developed and implemented three distinct control strategies tailored to meet the diverse needs of researchers. The first strategy enables direct control over each propeller, providing granular manipulation ideal for those delving into the complexities of advanced boat control and dynamics. The second strategy elevates the level of abstraction, allowing researchers to specify the USV's velocity and heading, thus catering to studies focused on navigation and dynamic obstacle avoidance without engaging with the details of the vessel's dynamics. The third strategy hides all the dynamic elements, concentrating solely on high-level strategic tasks such as bathymetry data collection and processing and static obstacle avoidance. This tiered approach ensures that our platform can support a wide range of research objectives, facilitating interdisciplinary studies and broadening the scope of potential applications.

By open-sourcing our platform and providing comprehensive documentation in the accompanying GitHub repository [63], we aim to empower the research community to build upon our work and drive innovation in bathymetry USV technology.

## ACKNOWLEDGMENT

This work was supported by the California Department of Transportation (CALTRANS). We extend our appreciation to Hrushikesh Mathi, Kokul Aananth Kathiravan Kavitha, and James Chow who assisted with testing, validation and debugging. The satellite basemap used in this work was made possible by Esri's World Imagery service, with imagery provided by Esri, Maxar, Earthstar Geographics, and the GIS User Community.

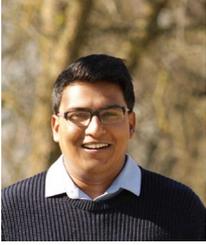

**DINESH KUMAR** Dinesh Kumar received his B.Tech. in Aerospace Engineering from Amrita Vishwa Vidyapeetham, Coimbatore, India, in 2020, and his M.S. in Mechanical and Aerospace Engineering from the University of California, Davis, USA, in 2023. He has experience in developing autonomous systems across both academia and industry. Since 2024, he has been a Research and Development Engineer at the Advanced Highway Maintenance and Construction Technology Research Center, University of California, Davis, where he focuses on hardware-software integration, mechanical design, prototyping, and system validation for UAVs and USVs.

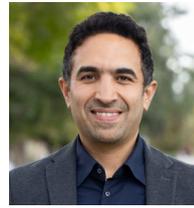

 **IMAN SOLTANI** is an assistant professor of Mechanical and Aerospace Engineering and a faculty member in graduate groups of the Departments of Computer Science and Electrical and Computer Engineering at the University of California, Davis. His research spans the interface of artificial intelligence, instrumentation, controls, and design, with a focus on machine learning and robotic systems. Before joining UC Davis, he worked at the Ford Greenfield Labs in Palo Alto, CA, where he founded and led the Advanced Automation Laboratory. He earned his bachelor's, master's, and PhD in mechanical engineering from Tehran Polytechnic (Iran), the University of Ottawa (Canada), and the Massachusetts Institute of Technology (MIT), respectively. He holds more than 18 patents and has authored over 40 journal and conference publications on topics ranging from medical imaging to autonomous driving, high-speed nanorobotics, dexterous bimanual robotics, machinery health monitoring, and precision positioning systems. His research has been featured in prominent outlets such as The Boston Globe, Elsevier Materials Today, ScienceDaily, and MIT News. Among his numerous awards are the MIT Carl G. Sontheimer Award and National Instruments' Engineering Impact Awar

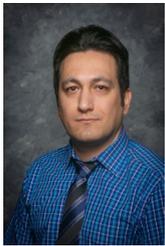

**AMIN GHORBANPOUR** earned his B.S. (2009) and M.S. (2011) in Aerospace Engineering from Amirkabir University of Technology and Tarbiat Modares University, respectively, in Tehran, Iran. He obtained his Ph.D. in Mechanical Engineering from Cleveland State University, Cleveland, OH, USA, in 2021 and is now an Assistant Professor of Mechanical Engineering at Duquesne University, Pittsburgh, PA, USA. His research focuses on control, robotics, and mechatronics, with applications in autonomous systems, aerospace guidance and navigation, and energy harvesting in mechanical systems.

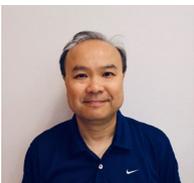

**KIN YEN** earned his bachelor's degree in mechanical engineering from the California State University at Fresno, California, US in 1994. He completed his master's degree in mechanical engineering from the University of California-Davis in 2002. He is a senior development engineer at the Advanced Highway Maintenance and Construction Technology Research Center, University of California, Davis since 2002. He has over 23 years of experience applying advanced automation, robotics, remote sensing, surveying, and related technologies for transportation system operations and maintenance. He has extensive project lead and participation experience with Caltrans projects including static and mobile terrestrial laser scanning systems (MTLS), bathymetric survey, Uncrewed Aerial System mapping, geographic information systems, driver assistance systems, and vision-based systems for automated site monitoring.